\documentclass[11pt,a4paper]{article}
\usepackage[hyperref]{emnlp2018}
\usepackage{times}
\usepackage{latexsym}
\usepackage{blindtext}
\usepackage{url}
\usepackage{amsmath}
\usepackage{amsfonts}
\usepackage{graphicx}
\usepackage{subcaption}
\usepackage{tabularx}

\aclfinalcopy 

\newcommand{\sugg}[1]{#1}

\title{Chargrid: Towards Understanding 2D Documents \\\mbox{ }}

\author{Anoop R Katti\thanks{~~Equal contribution} \\\And
  Christian Reisswig\footnotemark[1] \\\And
  Cordula Guder \\\And
  Sebastian Brarda \\\AND
  Steffen Bickel \\\And
  Johannes H\"ohne \\\\
  SAP SE\\
{\small Machine Learning R\&D, Berlin Germany } \\
{\scriptsize \{anoop.raveendra.katti $\vert$ christian.reisswig  $\vert$ cordula.guder $\vert$ sebastian.brarda $\vert$ steffen.bickel $\vert$ johannes.hoehne $\vert$ jean.baptiste.faddoul\}@sap.com}\And
Jean Baptiste Faddoul
}

\usepackage{fancyhdr}
\date{}
\def\c{\cite}

\begin{document}
\maketitle
\thispagestyle{fancy}
\begin{abstract}

We introduce a novel type of text representation that preserves the 2D layout of a document. 
This is achieved by encoding each document page as a two-dimensional grid of characters.
Based on this representation, we present a generic document understanding pipeline for structured documents. This pipeline makes use of a fully convolutional encoder-decoder network that predicts a segmentation mask and bounding boxes. We demonstrate its capabilities on an information extraction task from invoices and show that it significantly outperforms approaches based on sequential text or document images.

\end{abstract}

\section{Introduction}
\label{sec:introduction}
Textual information is often represented through structured documents which have an inherent 2D structure. This is even more so the case with the advent of new types of media and communications such as presentations, websites, blogs and formatted notebooks.
In such documents, the layout, positioning, and sizing might be crucial to understand its semantic content and provide a strong guidance to the human perception. 

NLP addresses the task of processing and understanding natural language texts through sub-tasks like language modeling, classification, information extraction, summarization, translation, and question answering among others. NLP methods typically operate on serialized text, which is a 1D sequence of characters. Such methods have been proven very successful for various tasks on unformatted text (e.g.~books, reviews, news articles, short text snippets).
However, when processing structured and formatted documents in which the relation between words is impacted not only by the sequential order, but also by the document layout, NLP can result in significant shortcomings.

Computer vision algorithms, on the other hand, are designed to exploit 2D information in the visual domain. Images are commonly processed with convolutional neural networks \c{lecun1998gradient, krizhevsky2012imagenet, DBLP:journals/corr/RenHG015, pinheiro2016learningRefineObjectSegments} (or likes) that preserve and exploit the 2D correlation between neighboring pixels. 
While it is feasible to convert structured documents into images and then apply computer vision algorithms, this approach is not optimal for understanding their semantics as it is driven mostly by the visual content and not by the textual content. As a result, a machine learning model would first need to extract the text from the image followed by learning the semantics. This purely visual approach requires a more complex model and significantly larger training data compared to text-based approaches.

We propose a novel paradigm for processing and understanding structured documents. Instead of serializing a document into a 1D text, the proposed method, named \emph{chargrid}, preserves the spatial structure of the document by representing it as a sparse 2D grid of characters. 
\sugg{We then formulate the document understanding task as instance-level semantic segmentation on \emph{chargrid}. More precisely, the model predicts a segmentation mask with pixel-level labels and object bounding boxes to group multiple instances of the same class.}
We apply the chargrid paradigm on an information extraction task from invoices and demonstrate that this method is superior to both, state-of-the-art NLP algorithms as well as computer vision algorithms.

The rest of the document is organized in three parts: we first introduce the \textit{chargrid} paradigm; we then describe a specific application of information extraction from documents; finally, we present experimental results and conclusion.

\section{Related Work}
\label{sec:relatedwork}

NLP focuses on understanding natural language text through tasks like classification \c{DBLP:journals/corr/Kim14f}, translation \c{DBLP:journals/corr/BahdanauCB14}, summarization \c{DBLP:journals/corr/RushCW15}, and named entity recognition \c{lample2016neural}. Such methods expect unformatted text as input and, therefore, do not assume any intrinsic 2D structure. 

Document Analysis, on the other hand, deals largely with problems such as recognizing printed/hand-written characters from a variety of documents \c{NIPS2008_3449}, processing document images for document localization \c{8269957}, binarization \c{abs-1708-03276}, and layout segmentation \c{7333914}. As a result, it does not focus on understanding the character- and/or word-level semantics of the document the same way as NLP.

Within computer vision, problems such as scene text detection and recognition \c{DBLP:journals/corr/GoodfellowBIAS13}, semantic segmentation \c{badrinarayanan2017segnet} as well as object detection \c{gupta2014learning, RetinaNet} can be considered as related problems to ours, but applied on a different domain, i.e., processing natural images instead of documents as input.

The closest to our work is  \citet{yang2017learning} that performs pixel-wise layout segmentation on a structured document, using sentence embeddings as additional input to an encoder-decoder network architecture. For each pixel inside the area of a sentence, the sentence embedding is appended to the visual feature embedding at the last layer of the decoder. The authors show that the layout segmentation accuracy can be significantly improved when using the textual features.
Another related work is \citet{palm2017cloudscan} which extracts key-value information from structured documents (invoices) using a recurrent neural network (RNN). Their work addresses the problem of document understanding, however, the RNN operates on serialized 1D text. 

Combining approaches from computer vision, NLP, and document analysis, our work is the first to systematically address the task of understanding 2D documents the same way as NLP while still retaining the 2D structure in structured documents. 

\section{Document Understanding with Chargrid}
A human observer comprehends a document by understanding the semantic content of characters, words, paragraphs, and layout components. We encapsulate all such tasks under the common umbrella of \textit{document understanding}. Therefore, we can formulate this problem as an instance segmentation task of characters on the page. In the following sections, we describe a new approach for solving that task. 

\subsection{Chargrid}
\label{sec:chargrid}
\textit{Chargrid} is a novel representation of a document that preserves its 2D layout.
A chargrid can be constructed from \textit{character boxes}, i.e.,~bounding boxes that each surround a single character somewhere on a given document page. This positional information can come from an optical character recognition (OCR) engine, or can be directly extracted from the layout information in the document as provided by, e.g., PDF or HTML. The coordinate space of a character box is defined by page height $H$ and width $W$, and is usually measured in units of pixels.

The complete text of a document page can thus be represented as a set of tuples $\mathcal{D} = \{(c_k, \mathbf{b}_k)\,|\,k=0,...,n\}$, where $c_k$ denotes the $k$-th character in the page and $\mathbf{b}_k$ the associated character box of the $k$-th character, which is formalized by the top-left pixel position, width and height, thus $\mathbf{b}_k = (x_k, y_k, w_k, h_k)$. 

We can now construct the chargrid $g\in \mathbb{N}^{H \times W}$ of the original document page, and its \textit{character-pixel} $g_{ij}$ from the set $\mathcal{D}$ with

\begin{equation}
\label{eq:chargrid}
g_{ij} = 
\begin{cases}
E(c_k) & \text{if}\quad (i,j)\prec\mathbf{b}_k \\
0 &
\end{cases}
\end{equation}
where $\prec$ means 'overlaps with', and
where each point $(i,j)$ corresponds to some pixel in the original document page pixel coordinate space defined by $(H,W)$. 
$E(c_k)$ is some encoding of the character in the $k$-th character box, i.e.~the value of character $c_k$ may be mapped to a specific integer index. For instance, we may map the alphabet (or any character of interest) to non-zero indices $\{a,b,c,...\} \rightarrow \{1,2,3,...\}$.
Note that we assume that character boxes cannot overlap, i.e.~each character on a document page occupies a unique region on the page. In practice, it may happen that the corners and edges of a character box may overlap with other closeby characters. We solve such corner cases by assigning the character-pixel to the box that has the closest box center.

\begin{figure}
  \begin{center}
    \includegraphics[width=0.5\textwidth]{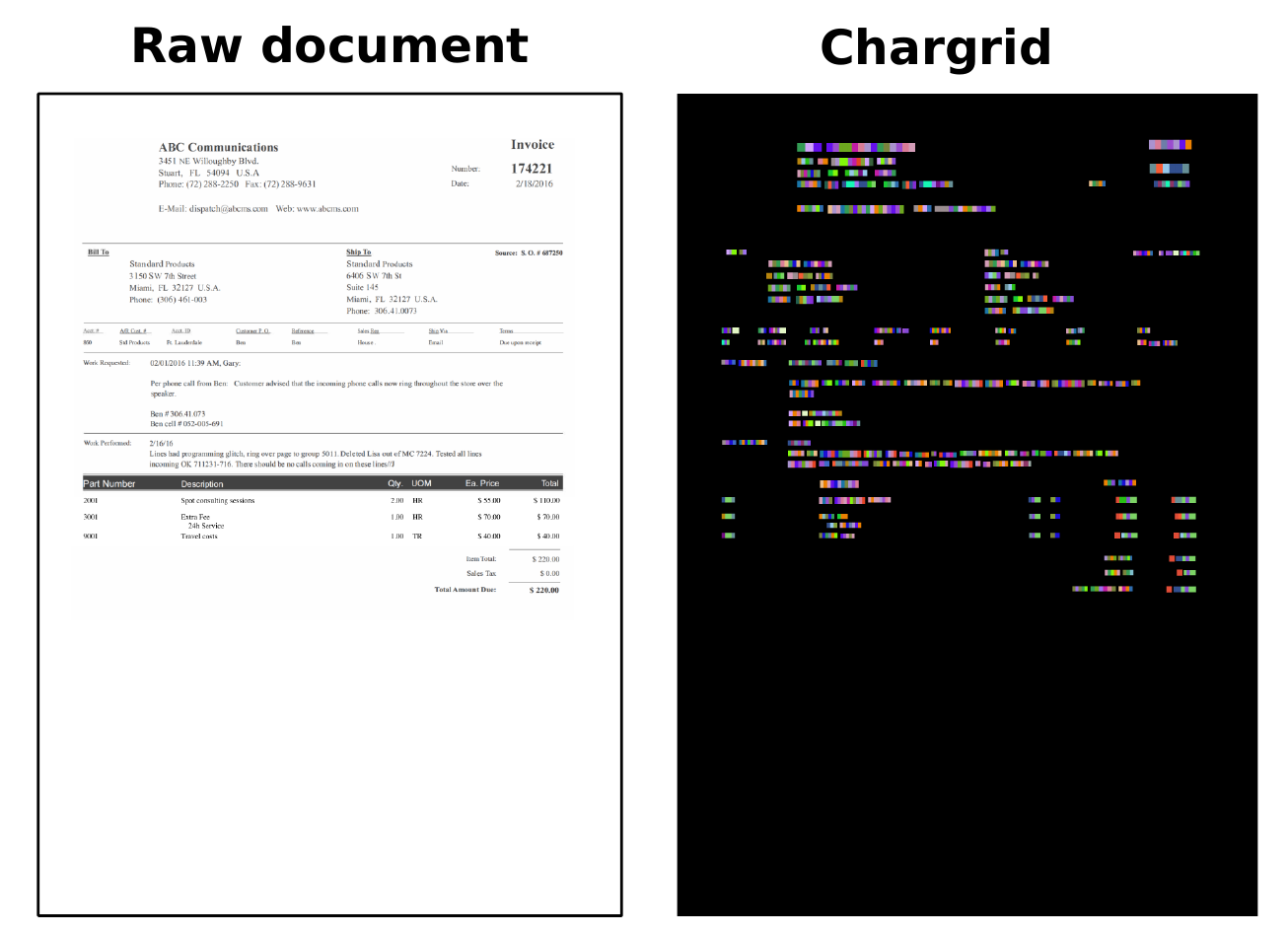}
  \end{center}
  \caption{Example for a document page (left) and corresponding chargrid representation $g$ (right).}
  \label{fig:chargrid}
\end{figure}

In other words, the chargrid representation is constructed as follows: for each character $c_k$ at location $\mathbf{b}_k$, the area covered by that character is filled with some constant index value $E(c_k)$. All remaining character-pixels corresponding to empty regions on the original document page are initialized with 0. Figure~\ref{fig:chargrid} visualizes the chargrid representation of an example input document.

The advantage of the new chargrid representation is twofold: (i) we directly encode a character by a single scalar value rather than by a granular collection of grayscale pixels as is the case for images, thus making it easy
for the subsequent document analysis algorithms to understand the document, and (ii), because the group of pixels that belonged to a given character are now all mapped to the same constant value, we can significantly downsample the chargrid representation without loss of any information.
For instance, if the smallest character occupied a $10\times10$ pixel region in the original document, we can downsample the chargrid representation by a factor of $10\times10$.
This significantly reduces the computational time of subsequent processing steps, such as training a machine learning model on this data representation.

We point out that a character can occupy a region spanning several character pixels. This is in contrast to traditional NLP where each character is represented by exactly one token. In our representation, therefore, the larger a given character, the more character-pixels represent that single character. 
We do not find this to be a problem, instead, it even helps since it implicitly encodes additional information, for example, the font size, that would otherwise not be available.

Before the chargrid representation is used as input to, e.g.,~a neural network (see Sec.~\ref{sec:networkarch}), we apply 1-hot encoding to the chargrid $g$. Thus, the original chargrid representation $g \in \mathbb{N}^{H \times W}$ becomes a vector representation $\tilde{g} \in \mathbb{R}^{H \times W \times N_C}$, where $N_C$ denotes the number of characters in the vocabulary including a padding/background character (in our case this is mapped to 0) and an unknown character (all characters that are not mapped by our encoding $E$ will be mapped to this character).

\sugg{We note that similar to using characters for constructing the chargrid, one can also use words to construct a \textit{wordgrid} in the same way. In that case, rather than using 1-hot encoding, one may use a word embedding like word2vec or GloVe. While the construction of a wordgrid seems straight-forward, we have not experimented with it in the present work as our dataset contains too many unusual words and spans multiple languages (see Sec.~\ref{sec:invoivoice_analysis}).}

\subsection{Network Architecture}
\label{sec:networkarch}
\begin{figure*}
  \begin{center}
    \includegraphics[width=1.\textwidth]{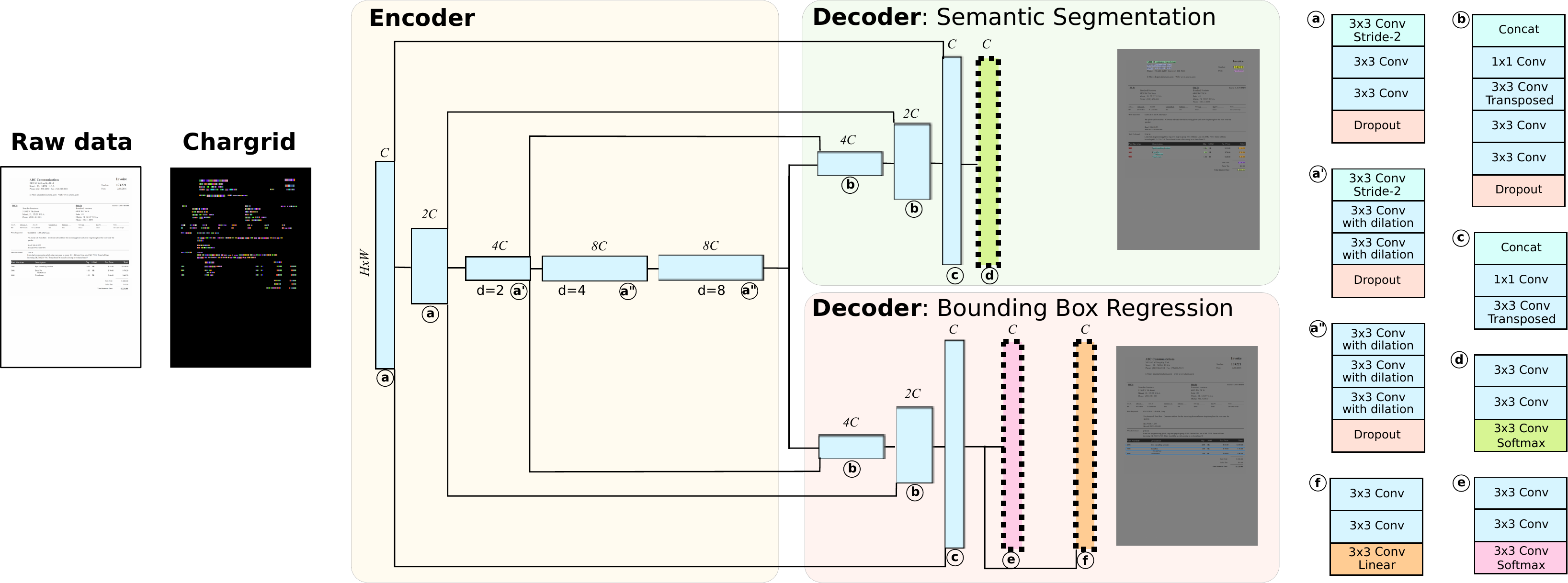}
  \end{center}
  \caption{Network architecture for document understanding, the \textit{chargrid-net}. Each convolutional block in the network is represented as a box. The height of a box is a proxy for feature map resolution while the width is a proxy for the number of output channels. $C$ corresponds to the number of 'base' channels, which in turns corresponds to the number of output channels in the first encoder block. $d$ denotes dilation rate.}
  \label{fig:networkarch}
\end{figure*}

We use the 1-hot encoded chargrid representation $\tilde{g}$ as input to a fully convolutional neural network to perform semantic segmentation on the chargrid and predict a class label for each character-pixel on the document. As there can be multiple and an unknown number of instances of the same class, we further perform instance segmentation. This means, in addition to predicting a segmentation mask, we may also predict bounding boxes using the techniques from object detection. This allows the model to assign characters from the same segmentation class to distinct instances.

Our model is described in Figure~\ref{fig:networkarch}. It is comprised of two main parts: The encoder network and the decoder network. The decoder network is further made up of two branches: The segmentation branch and the bounding box regression branch.
The encoder boils down to a VGG-type network \cite{simonyan2014very} with dilated convolutions \cite{YuKoltun2016}, batch normalization \cite{ioffe2015batch}, and spatial dropout \cite{tompson2015efficient}. Essentially, the encoder consists of five blocks where each block consists of three $3\times3$ convolutions (which themselves are made of convolution, batch normalization, ReLU activation) followed by spatial dropout at the end of a block. The first convolution in a block is a stride-2 convolution to downsample the input to that block. Whenever we downsample, we increase the number of output channels $C$ of each convolution by a factor of two. 
We have found stride-2 convolutions to yield slightly better results compared to max pooling. In block four and five of the encoder, we do not apply any downsampling, and we leave the number of channels at $512$ (the first block has $C=64$ channels). We use dilated convolutions in block three, four, five with rates $d=2,4,8$, respectively.

The decoder for semantic segmentation and for bounding box regression are both made of convolutional blocks which essentially reverse the downsampling of the encoder via stride-2 transposed convolutions.
Each block first concatenates features from the encoder via lateral connections followed by $1\times1$ convolutions \cite{UNET, pinheiro2016learningRefineObjectSegments}. Subsequently, we upsample via a $3\times3$ stride-2 transposed convolution. This is followed by two $3\times3$ convolutions. Note that whenever we upsample, we decrease the number of channels by a factor of two.

The two decoder branches are identical in architecture up to the last convolutional block. The decoder for semantic segmentation has an additional convolutional layer without batch normalization, but with bias and with softmax activation. The number of output channels of the last convolution corresponds to the number of classes.
Together with the encoder, the decoder for the bounding box regression task forms a one-stage detector which makes use of focal loss \cite{RetinaNet}. 
We also make use of the anchor box representation and corresponding bounding box regression targets as discussed in \citet{FasterRCNN}. The anchor-box representation allows us to handle bounding boxes that vary widely with respect to size and aspect ratios. Moreover, the anchor-box representation allows us to detect boxes of different classes.
The number of output channels are $2N_a$ for the box mask (foreground versus background) and $4N_a$ for the four box coordinates, where $N_a$ is the number of anchors per pixel.
The weights of all layers are initialized following \citet{DelvingDeepIntoRectifiers}, except for the last ones, which are initialized with a small constant value $1e-3$ for stabilization purposes. 

In total, we have three equally contributing loss terms: 
\begin{equation}
\mathcal{L}_{\mathrm{total}} = \mathcal{L}_{\mathrm{seg}} + \mathcal{L}_{\mathrm{boxmask}} + \mathcal{L}_{\mathrm{boxcoord}},
\end{equation}
where $\mathcal{L}_{\mathrm{seg}}$ is the cross entropy loss for segmentation, e.g.~\cite{UNET}, $\mathcal{L}_{\mathrm{boxmask}}$ is the binary cross entropy loss for box masks, and $\mathcal{L}_{\mathrm{boxcoord}}$ is the Huber loss for box coordinate regression \cite{FasterRCNN}. Both cross entropy loss terms are augmented following the focal loss idea \cite{RetinaNet}. We also make use of aggressive static class weighting in both cross entropy loss terms mainly to counter the strong class imbalance between irrelevant and "easy" pixels (the "background" class) versus relevant and "hard" pixels (all other classes; see Sec.~\ref{sec:impl-details} for further details).

We refer to the network depicted in Figure~\ref{fig:networkarch} as the \textit{chargrid-net}.

\section{Information Extraction from Invoices}
\label{sec:invoivoice_analysis}
As a concrete example for understanding structured 2D documents, we extract key-value information from invoices. We make no assumption on the format of the invoice, the country of origin (and consequently taxes, date formats, amount formats, currency etc.) or the language. In addition to that, real-world invoices often contain incomplete sentences, nouns, and abbreviations. 

We want our model to parse an invoice and to extract 5 \textit{header fields} (i.e., Invoice Number, Invoice Date, Invoice Amount, Vendor Name and Vendor Address) as well as the list of product items purchased, referred to as the \textit{line-items}. Line-items include details for each item such as Line-item Description, Line-item Quantity and Line-item Amount. Together with the background class, this yields 9 classes and each character on the invoice is associated to exactly one class. We note that while header fields may only appear once on an invoice (are unique), line-items may occur in multiple instances.

To extract the values for each field, we collect all characters that are classified as belonging to the corresponding class. For line-items, we further group the characters by the predicted item bounding boxes.

\subsection{Data}
\begin{figure}
    \centering
    \begin{subfigure}[t]{0.23\textwidth}
        \centering
        \includegraphics[width=\textwidth]{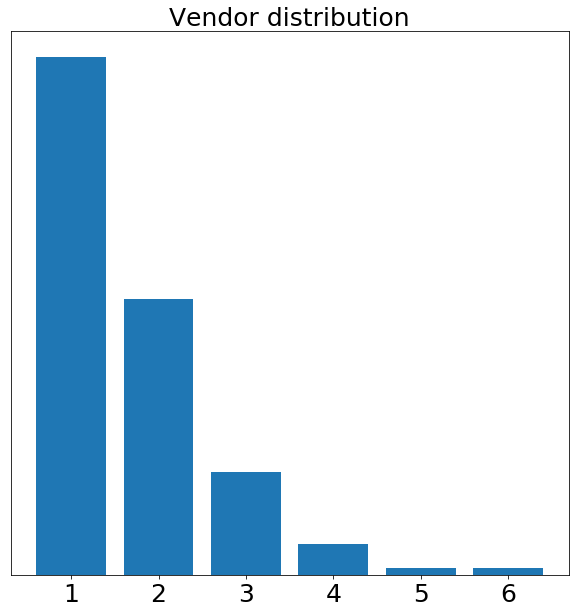}
    \end{subfigure}
    \begin{subfigure}[t]{0.23\textwidth}
        \centering
        \includegraphics[width=\textwidth]{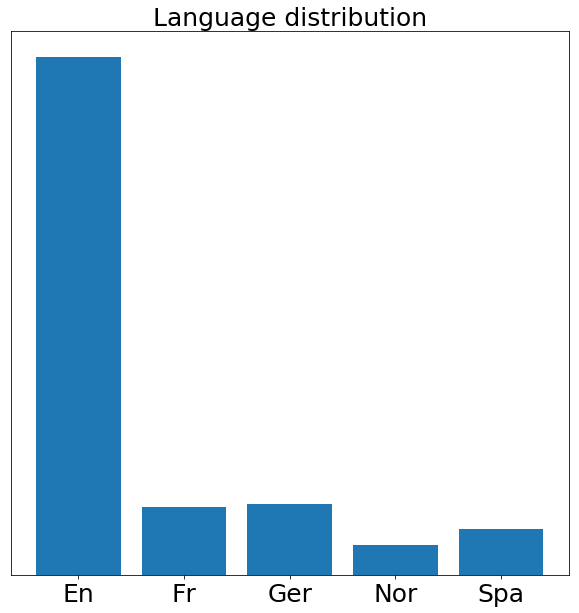}
    \end{subfigure}    
    \caption{The left image shows a histogram of number of vendors over the number of contributing invoices in the dataset. Most vendors appear only once in the dataset. The right image shows a distribution over languages, illustrating the diversity of the invoice data.}
   \label{fig:vendors_language}
\end{figure}
Our invoice dataset consists of 12k scanned sample invoices from a large variety of different vendors and languages. We assign 10k samples for training, 1k for validation and 1k for test on which we report our results (see Sec.~\ref{sec:results}). We ensure that vendors that appear in one set do not appear in any other set. This is more restrictive than necessary but gives a good estimate on how well the model generalizes to unseen invoice layouts. 
Figure~\ref{fig:vendors_language} shows the distribution over vendors and languages. From the vendor distribution, it can be seen that most vendors contribute only one invoice to the dataset with at most 6 invoices coming from a single vendor. From the language distribution, it can be seen that while English is the predominant language, there are large representations from French, Spanish, Norwegian, and German.

\begin{figure}

    \centering
    \begin{subfigure}[t]{0.23\textwidth}
        \centering
        \includegraphics[width=\textwidth]{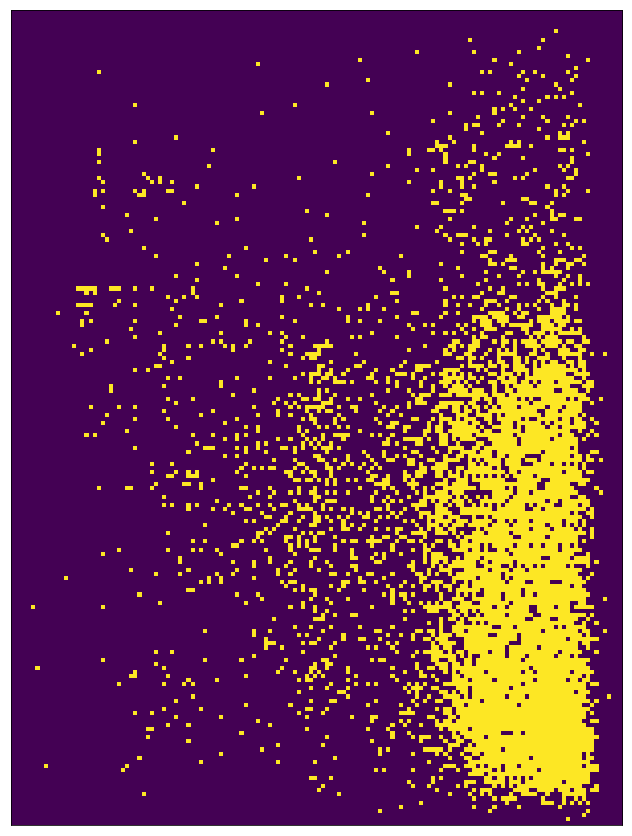}
    \end{subfigure}
    \hfill
    \begin{subfigure}[t]{0.23\textwidth}
        \centering
        \includegraphics[width=\textwidth]{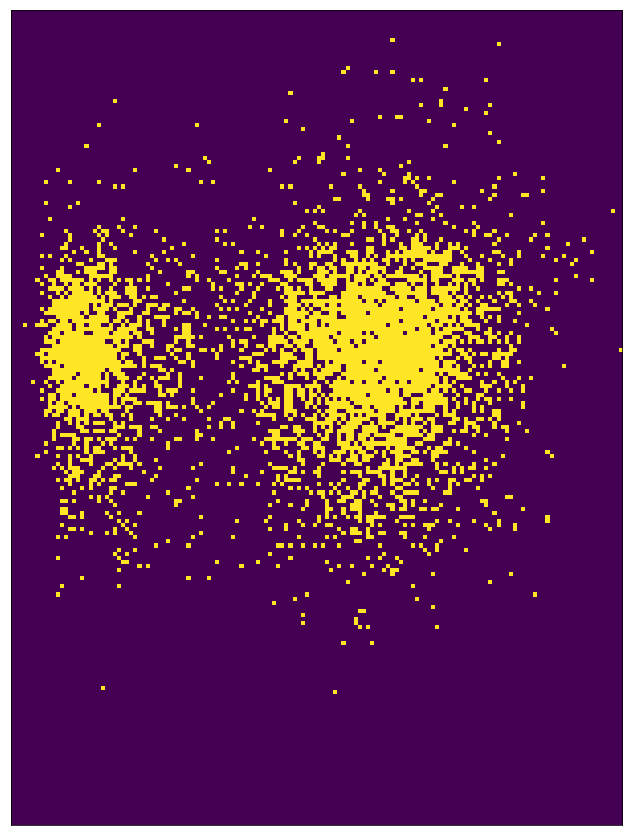}
    \end{subfigure}
    \caption{Spatial distribution of Invoice Amount (left) and Line-item Quantity (right) over the invoice. This depicts the variation in the invoice layouts contained in our dataset.}
    \label{fig:spatial_distribution}
\end{figure}

For all invoices, we collected manual annotations with bounding boxes around the fields of interest. Considerable efforts were spent to ensure that the labels are correct. In particular, each invoice was analyzed by three annotators plus a reviewer. Over 35k invoices were investigated and finally only clean set of 12k invoices were selected. Furthermore, detailed instructions were given to the annotators for each field. Figure~\ref{fig:spatial_distribution}  visualizes the locations of annotated boxes on invoices for Invoice Amount and Line-item Quantity. It can be seen that while some regions are denser, the occurrences are spread widely over the whole invoice thereby illustrating the diversity in the invoice format. 

\subsection{Implementation Details}
\label{sec:impl-details}
The invoices were processed with an open-source engine for OCR, \citet{tesseract}, to extract the text from the documents.

We limit the number of distinct characters in our encoding $E$ (in Eq.~\ref{eq:chargrid}) to the most frequent ones, which in our present case means $N_C=54$ different characters including the background/padding and unknown character. Thus, the 1-hot encoded chargrid $\tilde{g}$ has $54$ channels.
Each page of an invoice is processed independently. 

Once we created the chargrid representation of a document page, we downsample to a fixed resolution using nearest neighbor interpolation
to ensure that all chargrid representations have the same resolution for training (and inference). Note that this downsampling operation is still performed in the token space (i.e.,~before 1-hot encoding).
To further ensure that no tokens in the chargrid representation are lost due to nearest neighbor interpolation, we downsample to twice the target resolution. This resolution is determined by the smallest characters in the training set.
A second downsampling step, now in the 1-hot encoding to the final target resolution of 336x256 in our case, is performed using bilinear interpolation and fed into the network. 
Note that we could have also first performed 1-hot encoding on the document and applied bilinear interpolation directly to the target resolution. Computationally, however, this two-stage downsampling is more efficient.

We handle landscape documents by simply squeezing all input pages into our target resolution using interpolation (similar to image re-sizing). We have found that this approach is not harmful and for simplicity stick to it.

Line-items can occur in an unknown number of distinct instances. Therefore, we require instance segmentation of characters on the document. To accomplish this, the model 
is trained to predict bounding boxes that span across the entire row of one instance of a line-item, while the segmentation mask classifies those characters belonging to given column classes (such as, e.g.,~Line-item Quantity, or Line-item Description) of that line-item instance. 

We implemented our model in TensorFlow 1.4. 
We use SGD with momentum $\beta=0.9$ and learning rate $\alpha=0.05$. We used weight decay of $\lambda=10^{-4}$, and spatial dropout with probability $P=0.1$. 
We perform random cropping of the chargrid for data augmentation (that is we pad by 16 character pixels in each direction after downsampling and then crop with a random offset in range $(16,16)$). 

We use aggressive class weighting in the cross-entropy loss for semantic segmentation and for the bounding box mask. We have found this to be more effective than the focal loss (which can bee seen as a form of dynamic class weighting). We implement class weighting following \cite{paszke2016enet} with a constant of $c=1.04$. 
In early stages of our experiments not yet using class weights, we noted poor performance on the bounding box regression task as well as the segmentation task. 

The distribution of line-items on invoices in our dataset reveals that around $50\%$ of all invoices only contain less than three line-items. We found that repeating those invoices with more than three line-items during training more often than those with only few line-items significantly boosts bounding box regression accuracy. With a mini batch size of $7$, each model took around $360k$ iterations and 3 days to fully converge on a single Nvidia Tesla V100. 

\subsection{Evaluation Measure}
\label{sec:metric}
For evaluating our model, we would like to measure how much work would be saved by using the extraction system, compared to performing the field extraction manually.
To capture this, we use a measure similar to the word error rate \cite{prabhavalkar2017minimum} used in speech recognition or translation tasks. 

For a given field, we count the number of insertions, deletions, and modifications of the predicted instances (pooled across the entire test set) to match the ground truth instances. Evaluations are made on the string level. We compute this measure as follows:
\[1 - \frac{\#[\text{\footnotesize insertions}]+\#[\text{\footnotesize deletions}]+\#[\text{\footnotesize modifications}]}{N}\]
where $N$ is the total number of instances occurring in the ground truth of the entire test set.
This measure can be negative, meaning that it would be less work to perform the extraction manually. In our present case, the error caused by the OCR engine does not affect this measure, because the same errors are present in the prediction and in the ground truth and are not considered a mismatch.

\begin{figure*}
  \begin{center}
    \includegraphics[width=1\textwidth]{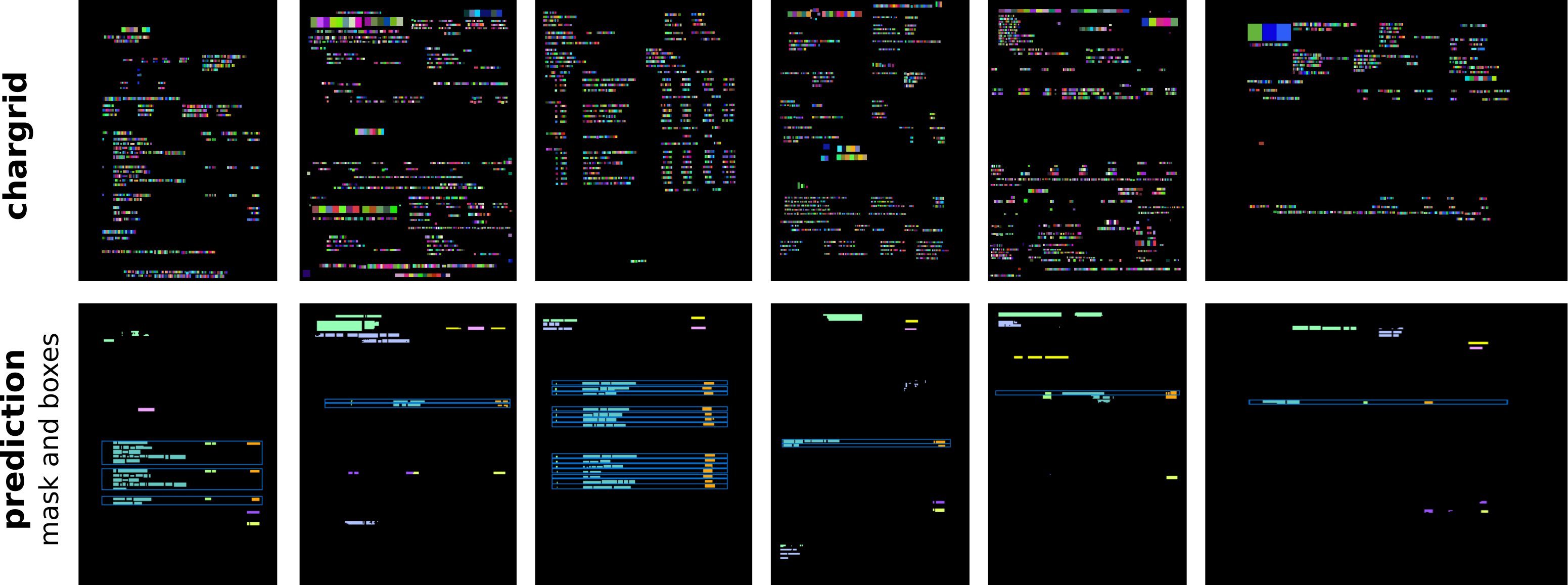}
  \end{center}
  \caption{Sample invoices and their corresponding network predictions: the top row shows the chargrid input, the bottom row shows the predicted segmentation mask with overlaid bounding box predictions. Our model is able to handle a large diversity of layouts. The encoding of the characters on the chargrid has been scrambled to preserve privacy.}
\label{fig:sampleinvoices}
\end{figure*}

\begin{figure}
  \begin{center}
    \includegraphics[width=0.48\textwidth]{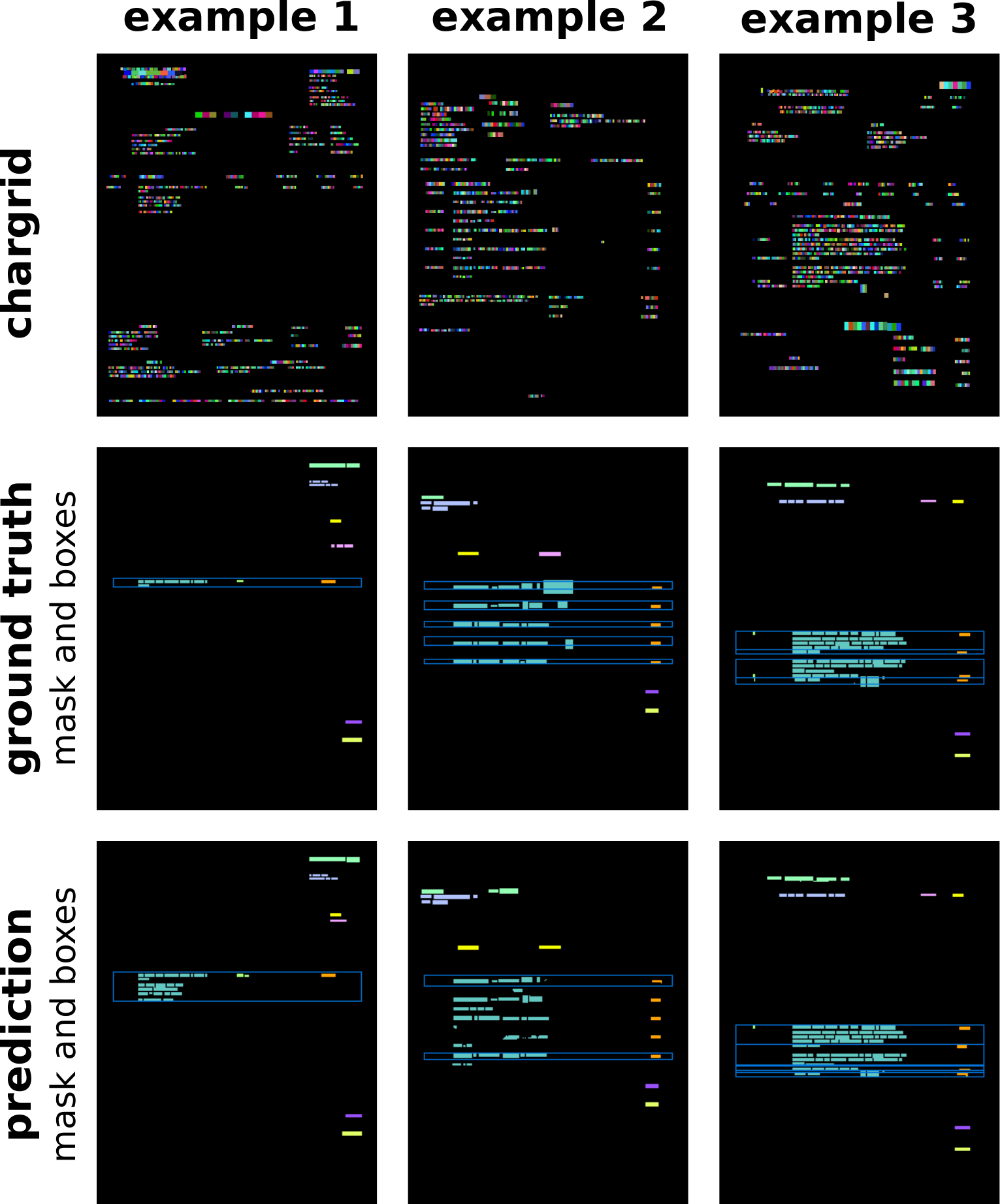}
  \end{center}
  \caption{\sugg{In example 1, chargrid-net attributes unrelated rows of text to the line-item's description. Moreover, some character-pixels are mis-classified such that the predicted header field above the line-item is incorrect. Thus, we observe errors in both, the segmentation mask and the bounding box predictions. In example 2, the model predicts the segmentation mask mostly correct, but it fails to predict boxes for two of the four line-items. In example 3, the model fails to separate adjacent multi-row line-items from each other correctly. Please note that the ground truth is debatable in some cases; for example, it may be hard to decide which information belongs to a line-item's description and which does not.}}
\label{fig:errorexamples}
\end{figure}

\section{Experiments and Results}
\label{sec:results}
Figure~\ref{fig:sampleinvoices} shows some sample predictions. It can be seen that the model can successfully extract key-value information on sample invoices despite significant diversity and complexity in the invoice layouts. 

We show the quantitative results in Table~\ref{tab:results}.
We compare the proposed model, chargrid-net (Sec.~\ref{sec:networkarch}), against four other models. The first one is a \textit{sequential} model based purely on text. It is a stack of bi-directional GRUs \c{DBLP:journals/corr/ChoMGBSB14} taking a sequence of characters as input, and producing a sequence of labels as output. This model is our implementation of \citet{palm2017cloudscan} and serves as a baseline comparison against a more traditional NLP approach which is based on sequential input. 
\sugg{We note that we have also experimented with a extension of this model that along with the characters also takes as input the position of each character on the document, however, with negligible returns. Therefore, we stick to this simpler model.}

The second \textit{image-only} model is identical to chargrid-net (Figure~\ref{fig:networkarch}), except that we directly take the original image of the document page as network input rather than the chargrid $\tilde{g}$. This model serves as a baseline comparison to a purely image-based approach using directly the raw pixel information as input. We note that we downsample the image to the same input resolution as the chargrid representation, that is 336x256.

The third and fourth models are both a \textit{hybrid} between the chargrid-net, and the image-only model. In both models, we replicate the encoder: one encoder for the chargrid input $\tilde{g}$, and one encoder for the image of the document page. Information from the two encoders is concatenated in the decoder: whenever a lateral connection from the encoder of the original chargrid-net is concatenated in a decoder block, we now concatenate lateral connections from the two encoders in a decoder block.

We distinguish two configurations of the hybrid model: model \textit{chargrid-hybrid-C64} where chargrid and image encoders both have $C=64$ base channels, and model \textit{chargrid-hybrid-C32} where chargrid and image encoders have $C=32$ base channels while the decoder still retains $C=64$ base channels. The latter model is used to assess the influence on the number of encoder parameters since the hybrid model effectively has more parameters due to the two encoders.

It can be seen that compared to the purely text-based approach, the chargrid-net performs equivalently on 
single-instance single-word fields where the 2D structure is not as important. Examples of single-instance, single-word fields are 'Invoice Number', 'Invoice Amount', and 'Invoice Date'. The extraction of these fields could essentially also be easily tackled with named entity recognition approaches based on serialized text \cite{gillick2015multilingual, lample2016neural}.

\begin{center}
\begin{table*}[htb]

{\small
\hfill{}
\begin{tabular}{|l|l|c|c|c|c|c|c|c|}
\hline
    Model/Field & \vtop{\hbox{\strut Invoice}\hbox{\strut Number}} & \vtop{\hbox{\strut Invoice}\hbox{\strut Amount}} & \vtop{\hbox{\strut Invoice}\hbox{\strut Date}} & \vtop{\hbox{\strut Vendor}\hbox{\strut Name}} & \vtop{\hbox{\strut Vendor}\hbox{\strut Address}} & \vtop{\hbox{\strut Line-item}\hbox{\strut Description}} & \vtop{\hbox{\strut Line-item}\hbox{\strut Quantity}} & \vtop{\hbox{\strut Line-item}\hbox{\strut Amount}} \\ \hline
    sequential & 80.98\% & 79.13\% & 83.98\% & 28.97\% & 16.94\% & -0.01\% & -0.18\% & 0.22\% \\ \hline
    image-only    & 47.79\% & 68.91\% & 45.67\% & 19.68\% & 13.99\% & 49.50\% & 46.79\% & 63.49\% \\ \hline
    chargrid-net & 80.48\% & 80.74\% & 83.78\% & 36.00\%& 39.13\%& 52.80\% & 65.20\% & 65.57\% \\ \hline
    chargrid-hybrid-C32  & 74.85\% & 77.93\% & 80.40\% & 32.00\% & 31.48\% & 46.27\% & 64.04\% & 63.25\% \\ \hline
    chargrid-hybrid-C64  & 82.49\% & 80.14\% & 84.28\% & 34.27\% & 36.83\% & 48.81\% & 64.59\% & 64.53\% \\ \hline
\end{tabular}
\hfill{}
\caption{Accuracy measure (c.f.~Sec.~\ref{sec:metric}) for an 8-class information extraction problem on invoices. The proposed chargrid models perform consistently well on all extracted fields compared to sequential and image models.}
\label{tab:results}
}\end{table*}
\end{center}

The chargrid-net, however, significantly outperforms the sequential model on multi-instance or multi-word fields where 2D relationships between text entities are important.
Examples of such fields are Line-item Description, Line-item Quantity and Line-item Amount, which are all grouped as \textit{sub-fields} to a specific line-item. Each of those fields may span a varying number of rows per line-item, see Figure~\ref{fig:sampleinvoices}.
The sequential model fails to correctly identify those line-item fields which is manifested in a negative accuracy measure (Sec.~\ref{sec:metric}). This implies that it is better to perform manual extraction over using automatic extraction. This is understandable since the line-item fields have a strong 2D structure that the sequential approach is not designed to capture.

In comparison with the image-only model, chargrid-net still performs much better. This is especially true for smaller fields which need to be \emph{read} to be accurately localized. On the other hand, for larger fields like Line-item Description, which can be localized by only vision, the gap is much smaller.

The values for the hybrid models are a bit more interesting. One could expect that combining two complementary inputs such as the chargrid representation and image - one capturing the content and the other capturing, e.g.~table delineations - would boost the accuracy. 
It turns out, however, that at least in the case of our invoice dataset, the additional image encoder does not bring additional benefits.
Model chargrid-hybrid-C64, where the chargrid encoder branch has the same number of base channels $C=64$ as the original chargrid-net, is essentially as accurate as chargrid-net.
Model chargrid-hybrid-C32, where the image and chargrid encoder combined have the same number of channels (that is the chargrid encoder branch only has $C=32$ base channels), the accuracy is significantly reduced. 

We conclude that at least in our present extraction problem, most of the discriminative information comes from the chargrid encoder branch and thus from the chargrid representation.

\sugg{Error examples of the chargrid-net model are depicted in Figure~\ref{fig:errorexamples}. One frequent error-type is that the model may fail to disentangle line-items, if they have a peculiar structure. While this is also challenging for human experts, other erroneous predictions are observed in samples for which the ground truth annotations are debatable.}

\section{Discussion}

We introduce a new way of modeling documents by using a character grid as document representation. The chargrid allows models to capture 2D relationships between characters, words, and larger units of text. The idea of the chargrid paradigm is inspired by the human perception which is heavily guided by 2D shapes and structures for understanding this type of documents.
Therefore, the chargrid allows to encode the positioning, size and alignment for textual components in a meaningful manner.
While the chargrid paradigm could be applied to various kinds of NLP tasks, we demonstrate its potential on an information extraction task from invoices. We train a deep neural network with an encoder-decoder architecture and we show that the network computes accurate segmentation masks and bounding boxes, which pinpoint the relevant information on the invoice.

We evaluate the accuracy of the model and compare it to state-of-the-art NLP and computer vision approaches. While those baseline models achieve accurate predictions for individual fields, only the chargrid performs well on all information extraction tasks. Some fields such as Invoice Number are relatively easy to detect for a model that operates on serialized text, as discriminative keywords are commonly preceding the words to be extracted. However, more complex extraction tasks (e.g. Vendor Address or Line-item Quantity) cannot be performed accurately as they require to exploit both, textual components and 2D layout structures. The traditional image-only computer vision model yields accurate predictions only for large visual columns (such as Line-item Amount) and it fails to locate extractions that require to understand the text. 

\sugg{However, compared to traditional sequential neural NLP models, the benefits in accuracy come at a larger computational cost compared. Even though the proposed model is fully convolutional and parallelizes very well on a single (or even multiple) GPU(s), the added complexity introduced by using a 2D data representation can significantly increase the total data dimensionality. In our current use case, a chargrid-net training requires up to three days until full convergence, whereas our sequential model converges after only a few hours.}

Comparing chargrid to semantic segmentation on natural images, one should note that character-pixels, unlike pixels of a gray-scale image, are categorical. This requires the character-pixels to be encoded as 1-hot. This yields a highly sparse data representation. While such representation is very common for NLP problems, it is new for segmentation networks that have previously only been applied to image segmentation.

\section{Conclusion}
\sugg{Chargrid is a generic representation for 2D text. Using this as a base, one could solve any task on a 2D text such as document classification, named-entity recognition, information extraction, parts-of-speech tagging etc. Furthermore, chargrid is highly beneficial for scenarios where text and natural images are blended. One could imagine performing all the above NLP tasks also on such inputs. This work demonstrates the advantages of chargrid for an information extraction task but we believe that it is only the first step towards incorporating the 2D document structure into document understanding tasks. Some follow-up direction could be solving other NLP tasks on structured documents using chargrid or experimenting with other computer vision algorithms on chargrid. Furthermore, it may be interesting to use word embeddings rather than 1-hot encoded characters, i.e.~a wordgrid, as 2D text input representation.}

\section*{Acknowledgments}
In acknowledgement of the partnership between Nvidia and SAP this publication has been prepared using Nvidia DGX-1. We thank Zbigniew Jerzak and Markus Noga for their support and fruitful discussions.

\bibliography{emnlp2018}
\bibliographystyle{emnlp_natbib}

\end{document}